# Deployment of an Agri-Bot for Greenhouse Administration

Ruchita R. Bhadre  
*Dept. of Instrumentation and Control*  
*College of Engineering, Pune*  
Pune, India  
ruchitabhadre@gmail.com

Prathamesh M. Yeole  
*Dept. of Instrumentation and Control*  
*College of Engineering, Pune*  
Pune, India  
yeolepm17.instru@coep.ac.in

*Abstract*—Modern agriculture is constantly evolving to increase production despite unfavorable environmental conditions. A promising approach is 'greenhouse cultivation' providing a microclimate to the cultivated plants to overcome unfavorable climate. However, massive-sized greenhouses develop non-uniform micro-climate throughout the complex requiring high degree of human supervision. We propose deploying an Agri-Bot to create and maintain positive ecological conditions in the greenhouse, reducing labor costs and increasing production. The prototype will contain two primary systems, the navigation system and the data analytics system. The navigation system will be controlled by an Arduino, and data analytics will be handled using an ESP8266 microchip. Numerous sensors for measuring the greenhouse parameters will be mounted on the robot. It will follow a predefined path, while taking readings at checkpoints. The microchip will collect and process data from sensors, transmit to the cloud, and give commands to the actuators. The soil and climate parameters like temperature, humidity, light intensity, soil moisture, pH will be measured periodically. When the parameters are not within a specified range, the Agri-Bot will take corrective actions like switching on blowers/heaters, starting irrigation etc. If external intervention is required, eg., fertilizer, it will indicate accordingly. Deploying such an Agri-Bot for monitoring and controlling microclimate in large-scale greenhouses can mitigate labor costs while increasing productivity. In spite of an initial cost, it can provide a high return on investment by providing flexibility, low power consumption and easy management to help greenhouse be water efficient, provide evenly dispersed and controlled sunlight intensity, temperature and humidity.

## I. INTRODUCTION

The major hurdle for productive yield in agriculture is the lack of skilled labor, limitations of human labor and insufficient data about farm environmental conditions. Due to insufficient data, farmers cannot predict required irrigation in the field due to which crops may not get balanced water supply. To avoid such detrimental situations, we should introduce an automated or notification system in the farm which would accelerate the process. Greenhouse cultivation is primarily useful in colder regions where normal open field cultivation is not feasible due to extreme weather conditions. Some crops such as tomato require very strict weather conditions, hence using a greenhouse would help in manipulating the environment inside the greenhouse for cultivation of such a crop. A greenhouse is specially used to develop positive styles of flowers at some stage in the 12 months or flowers that require non-stop tracking to gain excessive nice and quantity. At gift maximum of the greenhouses are manually managed and monitored. This approach of greenhouse tracking is labour extensive and time consuming. Therefore, we introduce an AGRI-BOT which is an agricultural robot that will replace human effort and also provide additional functionalities like regulating the environment inside the greenhouse.

The objective of this project is to develop a multipurpose robot platform for automated greenhouse to perform several tasks:

1. To maintain moisture content of soil.
2. To regulate temperature inside the greenhouse.
3. To maintain humidity inside the greenhouse.
4. Perform control actions for maintaining light intensity.
5. Give indications through notifications for the requirement of fertilisers.

## II. METHODOLOGY

Monitoring large greenhouses is much feasible solely by a human. To reduce human effort and improve performance and productivity, this job is done by the AGRI-BOT. The Wi-Fi controlled AGRI-BOT will traverse the specified path in the greenhouse. It will stop at marked checkpoints periodically and collect data with the help of sensors mounted on it. The data consists of values of various parameters in the greenhouse such as, temperature, humidity, Light intensity, soil moisture content, PH etc. This collected data will be processed in the microchip (NodeMCU). After processing the data, this information will be sent to the cloud. From the cloud the required information is received by the Wi-Fi module at the control end. From here control signals are given to the relays which will switch on the pump for water sprinklers or for fans and blowers as required. In this way with help of AGRI-BOT and Internet Of Things (IoT) corrective action for administration of Greenhouse is taken.

## III. HARDWARE AND SOFTWARE

### A. Power supply

The AGRI-BOT is powered by a Li-ION rechargeable battery. This battery is selected because :
- It has a long running time.
- It requires low maintenance.
- It has an internal protection circuit.

*B. Sensors*

*1) Temperature Sensor-LM35:* LM35 is a precession Integrated circuit Temperature sensor, whose output voltage varies, based on the temperature around it. It is a small and cheap IC which can be used to measure temperature anywhere between -55°C to 150°C. It can easily be interfaced with any Microcontroller that has ADC function or any development platform like Arduino. This sensor output is given to the digital pin of the Arduino and sensor is powered with the help of Arduino.

*2) Humidity Sensor-DHT11:* The DHT11 is a commonly used Temperature and humidity sensor. It uses a capacitive humidity sensor and a thermistor to measure the surrounding air, and spits out a digital signal on the data pin. This output data is given to the digital pin of ESP8266. This sensor is powered with the help of Arduino.

*3) Moisture Sensor-AR-65:* The AR-65 is a commonly used sensor to measure moisture of a soil. The main feature of the sensor is it gives digital output. It gives a digital output of 5V when the moisture level is high and 0V when the moisture level is low in the soil. The sensor includes a potentiometer to set the desired moisture threshold. When the sensor measures more moisture than the set threshold, the digital output goes high, and an LED indicates the output. When the moisture in the soil is less than the set threshold, the output remains low. The digital output of the sensor is connected to a digital pin of ESP8266.

*C. Relay JQC-3F*

Electrically operated switch is named as a relay. In Relay the current flowing via the coil creates a magnetic field which changes the switch contacts and attracts a lever. This relay is used to control pumps, fans, blowers with the help of ESP8266. The relay isolates the low power circuit from the high-power circuit and controls the high-power circuit with the help of low power circuit.

*D. Arduino Uno*

Arduino Uno is a microcontroller which is based on Atmel ATmega168 or ATmega328 and is frequently used in the area of automation. The operating voltage is 5V. However, the input voltage can vary between 7V-10V. IT has a flash memory of 32KB which is used to store the code. The Arduino comes with 14 digital and 8 analog pins which can be used to connect input devices such as sensors and output devices such as motor driver and relay circuit. The connections to pins of Arduino are shown in Fig.1.

*E. Motor Driver*

L293D is quadruple half H Bridge driver having a high current capability. L293D can provide currents up to 600mA and L293 can provide currents up to 1A at voltages ranging from 4.5 to 36. L293D is designed to drive inductive loads such as solenoids, relays, dc motors, bipolar stepper motors and other high current or voltage applications. It has 4 input pins which are connected to Arduino Nano and has 4 output pins which are connected to the dc motors. It uses the concept of H-bridge and hence can drive the dc motors in both the directions.

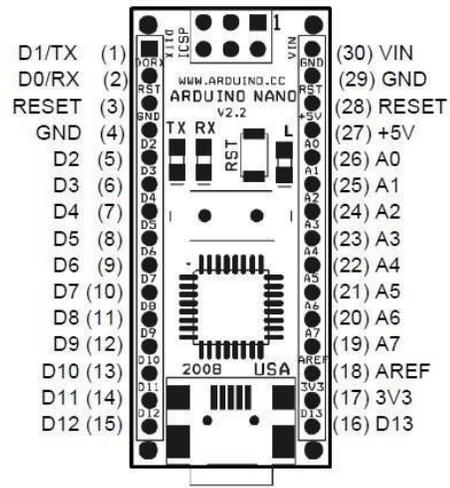

Fig. 1. Pin diagram of Arduino Uno.

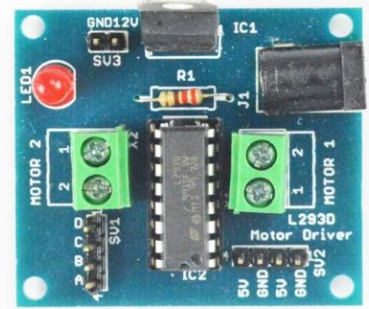

Fig. 2. LM293D Motor Driver board

*F. ESP8266*

NodeMCU is an open source IoT platform. It includes firmware which runs on the ESP8266 Wi- Fi SoC from Espressif Systems, and hardware which is based on the ESP12 module. ESP8266 is the Wi-Fi module that is interfaced to the Arduino through RxD and TxD pins. Thus, the Arduino through this Wi-Fi module when connected to a network can be controlled through a mobile application called Blynk. The working of the Blynk application is explained in Fig.4.

*G. Blynk application*

- Blynk is a Platform with iOS and Android apps to control Arduino over the internet.

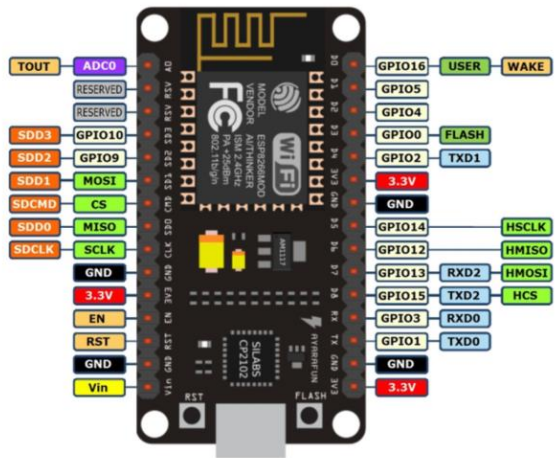

Fig. 3. NodeMCU

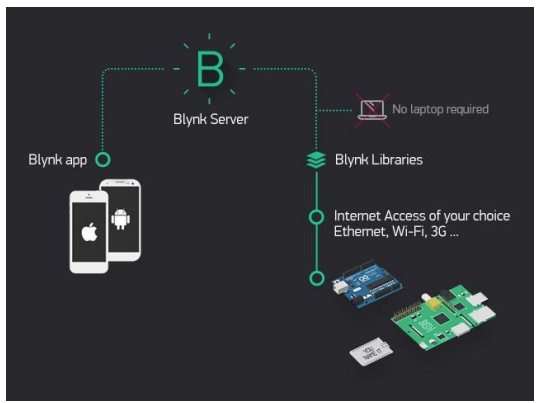

Fig. 4. Workflow of Blynk

- Operation of the bot is done through the mobile.
- In BLYNK app, the message travels to space the BLYNK Cloud, where it finds way to the connected hardware.

## IV. WORKING PRINCIPLE

### A. AGRI-BOT

*1) Navigation System:* This system handles the movement of Agri-Bot in a field. This system consists of ESP8266, motor driver, power supply, robot chassis and motor. Microchip receives a signal from the BLYNK software and according to the signal it controls the movement of the Agri-bot. The microchip is programmed in such a way that it enables the Agri-Bot to move in the farm automatically or manually. This system helps the farmer to drive the Agri-Bot by using the BLYNK software from anywhere in the world.

*2) Data Analytics System:* This system plays an important role in data acquisition with the help of different sensors. This system includes different sensors which are mounted on the Agri-Bot. With the help of navigation systems this system acquires the data from the checkpoints with the help of sensors mounted on Agri-Bot. The collected data will be processed in the ESP8266 microchip and then transmitted to the cloud. The data will be displayed on BLYNK software.

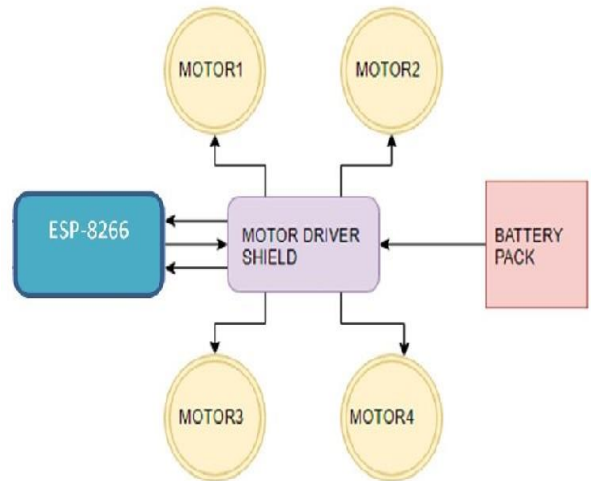

Fig. 5. Navigation system.

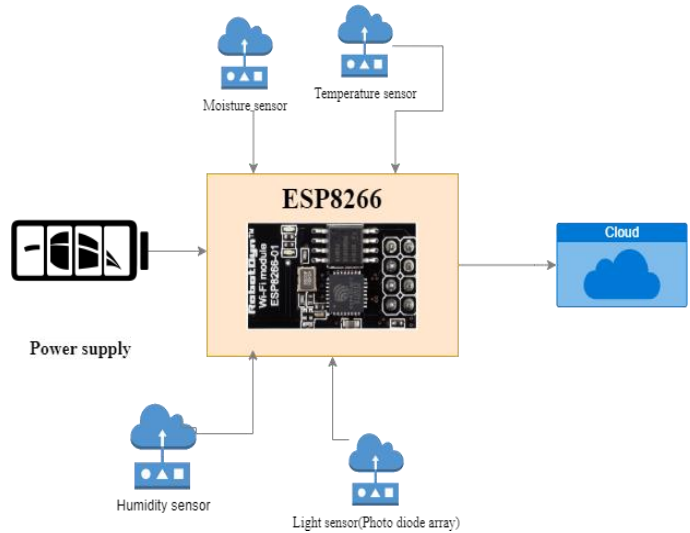

Fig. 6. Data Analytics System.

### B. Control System

This system involves the controlling action of actuators like fans, blowers etc. This system includes ESP8266, relay and actuator. In this system the ESP8266 microchip acquires data from the ESP8266 microchip which is mounted on the Agri-Bot with help of bridge communication between two ESP's. The soil and climate parameters like temperature, humidity, light intensity, soil moisture, pH will be measured periodically. When the parameters are not within a specified range, the Agri-

Bot will take corrective actions like switching on blowers/heaters, starting irrigation etc. with the help of relay.

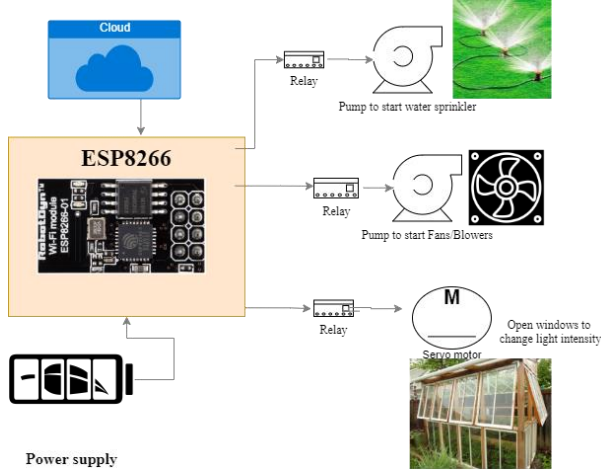

Fig. 7. Control system.

The data on cloud is processed and sent to ESP8226 for final control actions. The control system takes the final action based on the instruction given by ESP82266.

## V. RESULTS

The AGRI-BOT is completely assembled. The chassis for the AGRI-BOT is made from acrylic board. All the mentioned sensors are also mounted on the bot. Since the batteries and gear motor are heavy comparatively, normal wheels may not sustain the weight. Also, when moving on a moist soil, the wheels may be stuck inside the mud. To avoid this, larger wheels must be chosen. As the size of the AGRI-BOT increases, the rpm (rotations per minute) and power of the DC motors bus be chosen wisely to save malfunctioning of the device.

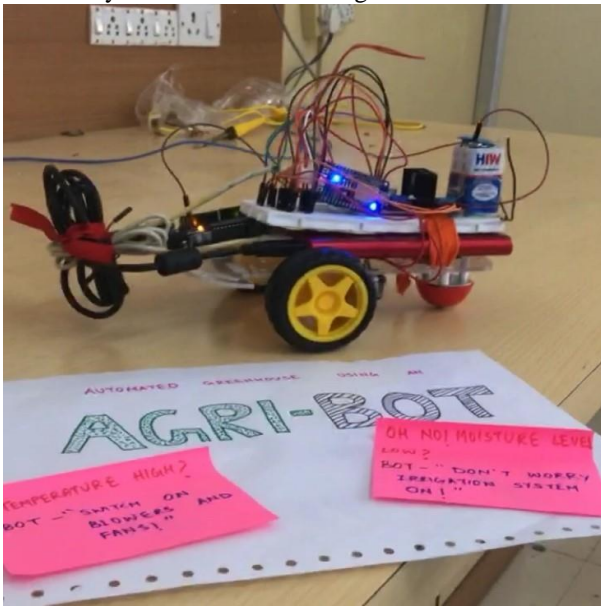

Fig. 8. Assembled AGRI-BOT

The Blynk Application is configured to control the AGRI-BOT's movements using the four virtual pins (LEFT, RIGHT, FORWARD and BACKWARD) as shown in the screenshot in Fig.9. The AGRI-BOT needs to be connected to a network in order to receive instructions from the application. The bot was placed in a virtual garden and tested.

The sensors are found to be working properly and their output value is displayed on the BLYNK Application as shown in the screenshot in Fig.9.

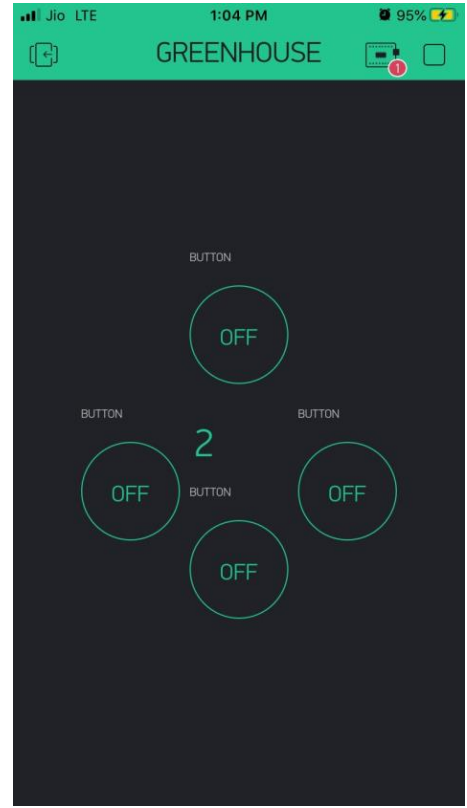

Fig. 9. Screenshot of Blynk App

## VI. CONCLUSION

The proposed model of AGRI-BOT for Greenhouse Management was developed. The AGRI-BOT is completely assembled according to the block diagram in Fig.5 and Fig.6 and tested. The movement is achieved through four dc shunt motors which are connected to motor driver The traditional system for greenhouse monitoring is labour-intensive and time consuming. The proposed system saves time, money and human effort. It provides a controlled environment for the plants to prevent them from damage and thus increasing the overall produce.

A few challenges were faced when the AGRI-BOT is placed on the field because of the weight of the batteries and the heavy DC gear motor. Another problem was sometimes the Bot might get stuck in the field. This was identified with the help of data received from the cloud. If the data was not changing for a long time indicated that the bot is stuck, and immediate human intervention is required in case of such emergencies.

ACKNOWLEDGMENT

With boundless love and appreciation, the authors would like to extend their heartfelt gratitude and appreciation to the people who helped us bring our study into reality. We would like to extend our profound gratitude to the Head of Department Dr. D. N. Sonawane, project mentors Dr. S. L. Patil and Mrs. Amruta Deshpande for their constant guidance and Dr. Abhishek Bihani from The University of Texas at Austin USA, for his unending support, advice and efforts to make this study possible.